\pdfoutput=1

\documentclass[11pt]{article}

\usepackage[final]{acl}

\usepackage{times}
\usepackage{latexsym}
\usepackage{amsmath}
\usepackage{amsfonts}

\usepackage[T1]{fontenc}

\usepackage[utf8]{inputenc}

\usepackage{microtype}

\usepackage{inconsolata}

\usepackage{graphicx}
\usepackage{multirow}
\usepackage{enumitem}

\usepackage{pifont}
\makeatletter
\renewcommand{\@fnsymbol}[1]{%
  \ifcase#1\or
    \ding{41}
  \else
    \@arabic{#1}
  \fi
}
\makeatother
\title{Graph-tree Fusion Model with Bidirectional Information Propagation for Long Document Classification}

\author{Sudipta Singha Roy$^{1}$, Xindi Wang$^{1,2}$\thanks{Corresponding Author.}, Robert E. Mercer$^{1}$, Frank Rudzicz$^{2,3,4}$\\
$^1$ Department of Computer Science, University of Western Ontario, Canada\\
$^2$ Vector Institute for Artificial Intelligence, Canada\\
$^3$ Faculty of Computer Science, Dalhousie University, Canada\\
$^4$ Department of Computer Science, University of Toronto, Canada\\
ssinghar@uwo.ca, xwang842@uwo.ca, mercer@csd.uwo.ca, frank@dal.ca}

\begin{document}
\maketitle
\begin{abstract}
Long document classification presents challenges in capturing both local and global dependencies due to their extensive content and complex structure. Existing methods often struggle with token limits and fail to adequately model hierarchical relationships within documents. To address these constraints, we propose a novel model leveraging a graph-tree structure. Our approach integrates syntax trees for sentence encodings and document graphs for document encodings, which capture fine-grained syntactic relationships and broader document contexts, respectively. We use Tree Transformers to generate sentence encodings, while a graph attention network models inter- and intra-sentence dependencies. During training, we implement bidirectional information propagation from word-to-sentence-to-document and vice versa, which enriches the contextual representation. Our proposed method enables a comprehensive understanding of content at all hierarchical levels and effectively handles arbitrarily long contexts without token limit constraints. Experimental results demonstrate the effectiveness of our approach in all types of long document classification tasks.
\end{abstract}

\section{Introduction}
Long document understanding has garnered increasing attention in the field of natural language processing (NLP) due to its wide range of applications across various domains, including legal document analysis, scientific literature categorization, and clinical text mining. Accurate understanding of long documents is essential for tasks such as information retrieval, content summarization, and decision-making support systems. Modern deep learning models for semantic analysis achieve impressive results by training on large datasets, which enables them to generate highly accurate predictions on unseen content~\cite{al-qurishi-2022-recent}. However, their ability to capture relationships between words and sentences relies on increasingly complex statistical operations as the text sequence lengthens~\cite{tay2020long}. Consequently, many existing methods become impractical for real-world applications, which makes processing long documents  a challenging task.
 
One of the primary challenges of long document classification is managing the large volume of information. Unlike short texts, long documents contain extensive content that often spans multiple topics, making it difficult to capture the overall context of the document effectively.
Transformer-based models~\cite{vaswani2017attention}, such as BERT~\cite{devlin-etal-2019-bert}, RoBERTa~\cite{liu2019roberta}, GPT-3~\cite{NEURIPS2020_1457c0d6} and LLaMa-2~\cite{touvron2023llama}, have  gained popularity for NLP tasks due to their ability to capture (relatively) long-range dependencies and contextual relationships. However, in long document classification, transformer models face scalability issues due to their quadratic time complexity with respect to the input length. Processing long documents with transformers can be computationally expensive and memory-intensive, often requiring substantial hardware resources. 
Current methods for handling lengthy documents include truncating texts to a predefined length or modifying the attention mechanism. Truncating to the first 512 tokens is straightforward but may cause significant information loss. Sparse attention models like Longformer~\cite{beltagy2020longformer} and Big Bird~\cite{10.5555/3495724.3497174} reduce computational load by focusing on a subset of tokens, but they do not fully capture comprehensive context and dependencies in long texts.

Another significant challenge in long document classification is capturing contextual dependencies and the hierarchical structure. Long documents have dependencies at word, sentence, and document levels, which are crucial for understanding content and leveraging both local and global contexts. 
This structure is important for understanding the overall context and meaning. Treating text as a flat sequence of tokens can cause models to miss these important hierarchical relationships.
Current approaches to address these challenges involve hierarchical models. For instance, Hierarchical Attention Network (HAN)~\cite{yang-etal-2016-hierarchical} and Hierarchical Attention Transformers (HAT) ~\cite{chalkidis-etal-2022-hat}, aim to capture both word-level and sentence-level representations before aggregating them into document-level embeddings. However, these models often fail to capture the intricate relationships between different parts of the document, such as the interplay between words, sentences, and overall document structure~\cite{dai-etal-2019-transformer}. 
Additionally, hierarchical models may struggle with long-range dependencies~\cite{dong2023survey}, which misses the relationships between distant sections essential for understanding the overall context.

To address the aforementioned challenges and constraints, we propose a novel model that leverages a graph-tree structure for arbitrarily long document classification. Our approach fuses syntax trees for sentence encodings with document graphs for document encodings, which provides a comprehensive representation that captures both local and global dependencies, respectively. Syntax trees (both dependency and constituency) represent the grammatical structure of sentences, which enhances sentence-level understanding. We  use Tree Transformers~\cite{ahmed-etal-2019-need} to generate sentence encodings from these syntax trees. The document graph preserves hierarchical relationships within the document, which ensures that both local and global contexts are considered during the classification process. We apply the Graph Attention Network (GAT)~\cite{velickovic2018graph} on the constructed document graph to model the dependencies between the sentences. This graph structure effectively captures both inter- and intra-sentence dependencies. Meanwhile, during training, we implement a bidirectional information propagation approach where information flows both from word-to-sentence-to-document and from document-to-sentence-to-word. This bidirectional flow enriches the contextual representation of the document, which allows for a more comprehensive understanding of the content at all hierarchical levels. By incorporating syntax trees and document graphs, we can encode text using different semantic units, such as sentence-level and document-level representations, based on their unique characteristics. This allows our model to handle arbitrarily long contexts without being constrained by token limits.

To summarize, our main contributions are:
\begin{enumerate}[noitemsep]
    \item We introduce a novel graph-tree structure that combines syntax trees and document graphs to  capture both local and global dependencies within arbitrarily long documents.
    \item We introduce a bidirectional information propagation approach where the information flows both from word-to-sentence-to-document and from document-to-sentence-to-word, which enriches the contextual representation of the document.
    \item We show empirically that our  model achieves improvements across a variety of classification tasks, including binary, multi-class, and multi-label classification. 
\end{enumerate}

\begin{figure*}[ht]
    \centering
    \includegraphics[width=\textwidth]{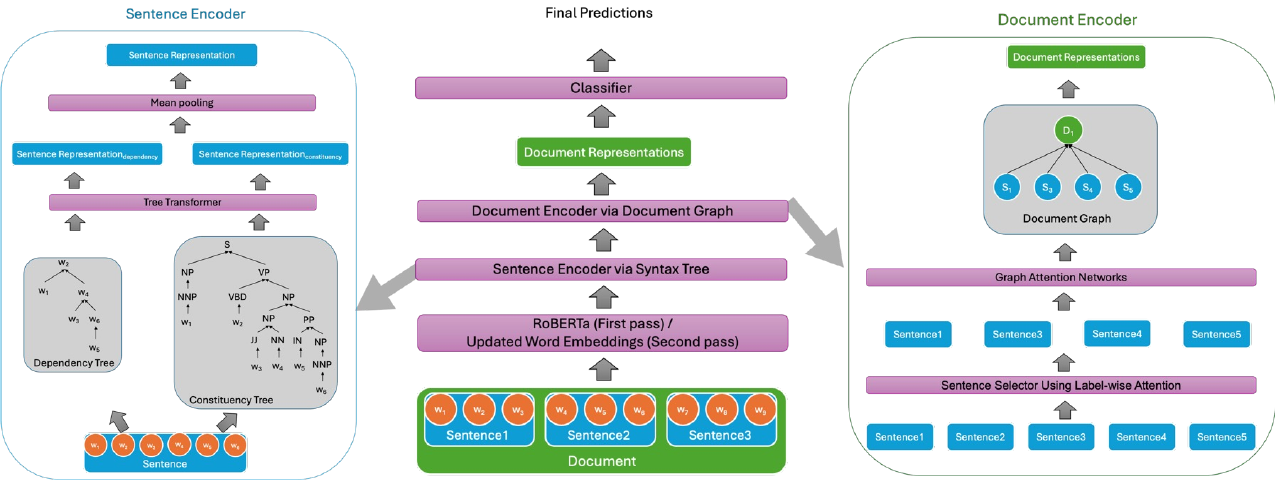}
    \caption{Overview of our proposed model. Our proposed model integrates syntax trees for sentence encodings (shown on the left) and document graphs for document encodings (shown on the right). We employ Tree Transformers to generate sentence encodings from the syntax trees and use graph attention networks to generate document encodings from the document graph. Our model workflow includes two passes. In the first pass, initial word embeddings are obtained from RoBERTa. In the second pass, these word embeddings are updated through bidirectional information propagation. }
    \label{fig:model}
\end{figure*}

\section{Preliminaries}
\subsection{Tree Transformer} \label{sec:tree}
Tree Transformer~\cite{ahmed-etal-2019-need} is designed to more effectively preserve syntactic and semantic information. Given a dependency or constituency tree structure of a sentence, a dependency tree has a word at every node, represented by $\mathbf{X}_d$ while, in a constituency tree, only the leaf nodes contain words, represented by  $\mathbf{X}_c$: 
\begin{equation} \label{eq:tree_rep}
    X_d = \begin{bmatrix}
\mathbf{p}_v \\
\mathbf{c}_1^v \\
\mathbf{c}_2^v \\
\vdots \\
\mathbf{c}_n^v
\end{bmatrix}, \quad
X_c = \begin{bmatrix}
\mathbf{c}_1^v \\
\mathbf{c}_2^v \\
\vdots \\
\mathbf{c}_n^v
\end{bmatrix},
\end{equation}
where $\mathbf{p}_v$ is the initial parent representation and $\mathbf{c}_i^v$ is the initial child representation of node $i$.

The parent node embedding $\mathbf{P}$ is computed using multi-branch attention built upon multi-head attention in the vanilla Transformer~\cite{vaswani2017attention}. 
The branch attention $\mathbf{B}_i$ for branch $i$ is computed as :
\begin{equation}
    \mathbf{B}_i = \mathrm{Attention}(\mathbf{Q}_i \mathbf{W}^Q_i, \mathbf{K}_i \mathbf{W}^K_i, \mathbf{V}_i \mathbf{W}^V_i),
\end{equation}
where $\mathbf{W}^Q_i$, $\mathbf{W}^K_i$, and $\mathbf{W}^V_i$ are the learnable weight matrices. 
Each $\mathbf{B}_i$ is then normalized and scaled using a layer normalization block~\cite{ba2016layer}:
\begin{equation}
    \overline{\mathbf{B}}_i = \mathrm{LayerNorm}(\mathbf{B}_i \mathbf{W}^b_i + \mathbf{B}_i) \times \kappa_i,
\end{equation}
where $\mathbf{W}^b_i$ and $\kappa_i$ are the learnable parameters. Then, a position-wise convolutional neural network (PCNN) is applied to each $\overline{\mathbf{B}}_i$, and the branch attention is aggregated:
\begin{equation}
\hspace{-2.23mm}
    \mathrm{BranchAttn}(\mathbf{Q}, \mathbf{K}, \mathbf{V}) = \sum_{i=1}^n \alpha_i \mathrm{PCNN}(\overline{\mathbf{B}}_i),
\end{equation}
where $\alpha_i$ is learnable. The final parent representation, or sentence embedding, is obtained by:
\begin{equation}
    \mathbf{P}' = \mathrm{EwS}(\tanh((\mathbf{x}' + \mathbf{x})\mathbf{W} + b)),
\end{equation}
where $\text{EwS}$ is element-wise summation, and $\mathbf{x}$ and $\mathbf{x}'$ depict the input and output of the attention module, respectively.
Additional details on the full operations of the Tree Transformer are in the original paper~\cite{ahmed-etal-2019-need}. 
\subsection{Graph Attention Network} \label{sec:gat}
Graph Attention Network (GAT)~\cite{velickovic2018graph} is designed to model information flow between nodes, which enhances node representations by employing  attention  over features from neighbouring nodes. Given a heterogeneous graph $G = (V, E)$ with $N$ nodes, 
the input node features $ h = \{ \mathbf{h}_1, \mathbf{h}_2, \dots, \mathbf{h}_N \} $ and GAT layer with multi-head attention are designed as follows:
\begin{equation}
    \begin{gathered}
    e_{ij} = \mathrm{LeakyReLU}\left(\mathbf{a}^T \left[\mathbf{W} \mathbf{h}_i \parallel \mathbf{W} \mathbf{h}_j \right] \right) \\
    \alpha_{ij} = \frac{\exp(e_{ij})}{\sum_{k \in \mathcal{N}_i} \exp(e_{ik})} \\
    \mathbf{h}_i' = \parallel_{k=1}^K \sigma\left( \sum_{j \in \mathcal{N}_i} \alpha_{ij}^k \mathbf{W}^k \mathbf{h}_j \right),
    \end{gathered}
\end{equation}
where $\parallel$ denotes concatenation, $\mathbf{W}$ is a weight matrix, $\mathbf{a}$ is a weight vector, $\mathcal{N}_i$ is the neighbourhood of node $i$, LeakyReLU and $\sigma$ are the activation functions, $K$ is the number of attention heads, $\alpha_{ij}^k$ and $\mathbf{W}^k$ are the attention coefficients and weight matrix for the $k^{th}$ head, respectively.

\section{Method}
We propose a novel Graph-Tree Fusion Model, as shown in Figure \ref{fig:model}
that leverages a graph-tree structure for long document classification. Our model uses multi-granularity document representations through Tree Transformers and graph attention networks. It fuses syntax trees for sentence encodings with document graphs for document encodings. Additionally, during training, we implement a bidirectional information propagation approach, allowing information to flow both from word-to-sentence-to-document and from document-to-sentence-to-word. 
\subsection{Sentence Encoder via Syntax Tree} \label{sentenceEncoder}
Sentences are foundational units of a document. Preserving sentence semantics enhances the accuracy and informativeness of document embeddings. 

We first split a document $\mathcal{D}$ into sentences as $\mathcal{S} = \{ s_1, s_2, \ldots, s_N\}$. Given a sentence $s$ with a sequence of input tokens, we use RoBERTa~\cite{liu2019roberta} to encode the tokens and output the corresponding vector for each token from the last hidden layer, denoted as $\mathbf{E}(s) = \{\mathbf{e}_1, \mathbf{e}_2, \ldots, \mathbf{e}_n\}$, where $\mathbf{e}_i$ is the embedding for token $i$. We then parse each sentence, obtaining an initialized dependency tree $\mathbf{X}_d^{s_i}$ and constituency tree $\mathbf{X}_c^{s_i}$ (see Equation \ref{eq:tree_rep}). 
These two tree structures are then processed by the Tree Transformers mentioned in Section~\ref{sec:tree}, yielding enhanced sentence representations for each sentence: $\mathbf{h}_d^{s_i}$ for the dependency tree and  $\mathbf{h}_c^{s_i}$ for the constituency tree:
\begin{equation}
    \begin{gathered}
        \mathbf{h}_d^{s_i} = \mathrm{TreeTransformer}(\mathbf{X}_d^{s_i}) \\
        \mathbf{h}_c^{s_i} = \mathrm{TreeTransformer}(\mathbf{X}_c^{s_i}). 
    \end{gathered}
\end{equation}
The final embedding for sentence $i$ is defined as the mean-pooling of the two tree representations:
\begin{equation}
    \mathbf{h}^{s_i} = \frac{1}{2}(\mathbf{h}_d^{s_i} + \mathbf{h}_c^{s_i}).
    \label{equ:meanPoolSentence}
\end{equation}
\subsection{Document Encoder via Document Graph} \label{documentEncoder}
\paragraph{Sentence Selector Using Label-wise Attention.}
To identify the most important sentences, we calculate the similarity score between each sentence $s_i$ in $\mathcal{D}$ and the labels using label-wise attention. We obtain the label embeddings for each label $\mathcal{Y}_i$, where $\mathcal{Y}_i$ is an element in the label set $\mathcal{Y}$. We calculate the label embeddings by taking the average of the word embeddings (from RoBERTa) for each word in their label names:
\begin{equation}
    \mathbf{h}^{\mathcal{Y}_i}= \frac{1}{m}\sum_{j \in m}w_{j}, i=1, 2, \ldots, L,
\end{equation}
where $m$ is the number of words in the label name, and $L$ is the number of labels. The similarity score is:
\begin{equation}
    \alpha_{s_{i}, \mathcal{Y}_i} = \mathrm{Softmax}(\mathbf{h}^{s_i} \cdot \mathbf{h}^{\mathcal{Y}_i}),
\end{equation}
where $\alpha_{s_{i}, \mathcal{Y}_i}$ is the probability score associated with sentence $i$ in $\mathcal{D}$ to a specific label $\mathcal{Y}_i$. Given the length of the document, considering all sentences with lower label-wise attention values would increase the computational burden. To address this, we apply a threshold 
$\tau$ to select only sentences with high probabilities:
\begin{equation} 
    \mathcal{S}' = \{s_i \vert \alpha_{s_{i}, \mathcal{Y}_i} \geq \tau\}.
\end{equation}
\paragraph{Document Encoding Using Graph Attention Network (GAT).} 
To obtain the document representation, we construct a heterogeneous document graph $G=(V,E)$ that captures the relations between documents and their sentences. The graph $G$ contains sentence nodes and document nodes, and one type of edge: document-sentence edges. Specifically, for each document $\mathcal{D}$, we create a document node $v_{\mathcal{D}}$ and a set of selected sentence nodes $V_{\mathcal{S}'} = \{v_{s'_{1}}, v_{s'_{2}}, ..., v_{s'_{K}}\}$. Directed edges $E$ are established from each sentence node $v_{s'_i}$ to its corresponding document node $v_{\mathcal{D}}$.

We apply GAT (described in \ref{sec:gat}) to model the inter-sentence and document relations within a document, which enhances the document representations derived from sentence representations.  

To obtain the document representation for $\mathcal{D}$, the sentence node embeddings $\mathbf{h}^{s'_i}$ are initially generated using the sentence encoder. The document node feature is then initialized by taking the mean pooling of the features of its sentence nodes:
\begin{equation}
\mathbf{h}^{\mathcal{D}} = \frac{1}{|{\mathcal{S}'}|} \sum_{{s'_i} \in {\mathcal{S}'}} \mathbf{h}^{s'_{i}}.
\end{equation}
We use the GAT layer with multi-head attention to compute the new document node features $\mathbf{h'}^{\mathcal{D}}$ as follows:
\begin{equation}
    \mathbf{h'}^{\mathcal{D}} = \mathrm{GAT}(\{\mathbf{h}^{s'_i}_{(i \in 1,2, ..., K)}, \mathbf{h}^{\mathcal{D}} \}).
\end{equation}
After the GAT layer, we introduce a position-wise feed-forward (FFN) layer, consisting of two linear transformations similar to the vanilla Transformer architecture~\cite{vaswani2017attention}, to obtain the final document representation:
\begin{equation}
    \tilde{\mathbf{h}}^{\mathcal{D}} = \mathrm{FFN}(\mathbf{h'}^{\mathcal{D}}).
    \label{eq:ffn}
\end{equation} 
\subsection{Bidirectional Information Propagation}
\begin{figure}[t]
\includegraphics[width=\columnwidth]{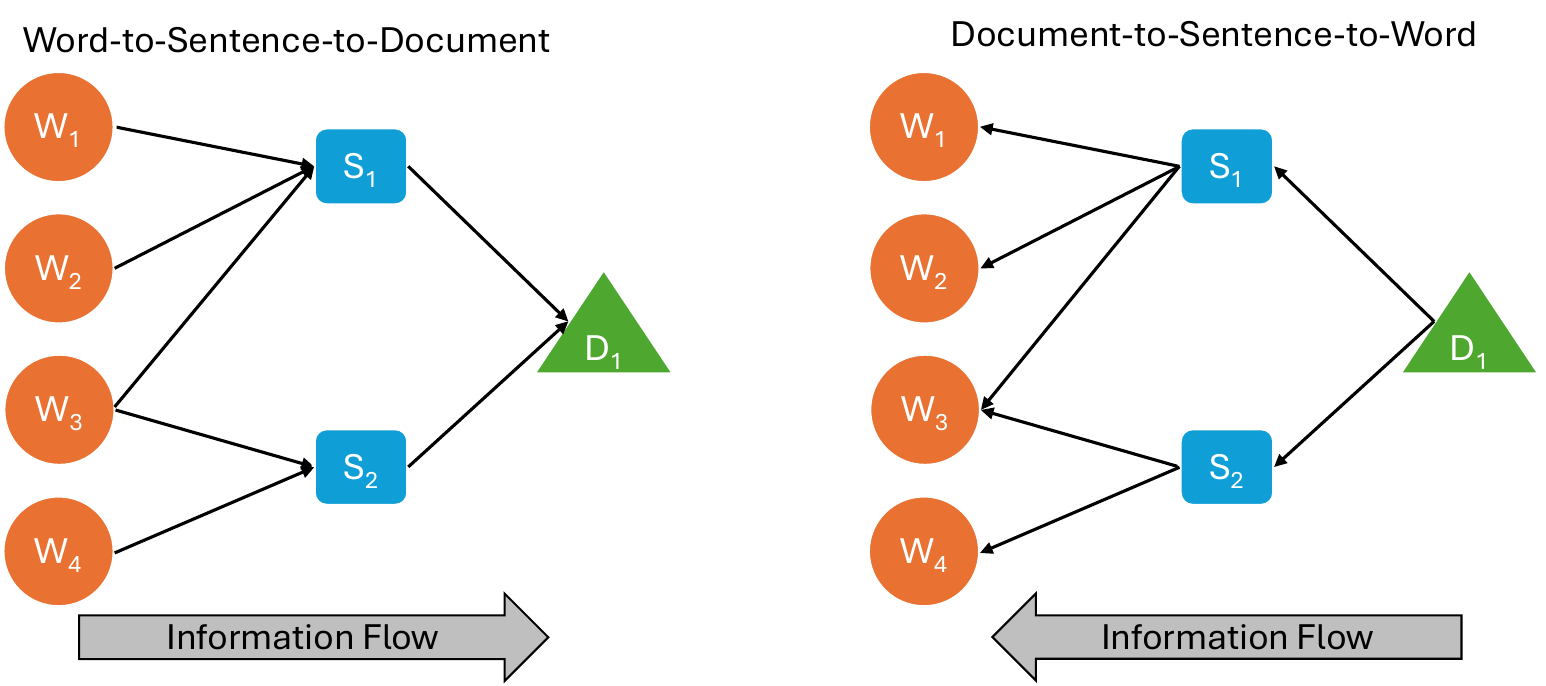}
\caption{The detailed bidirectional information propagation. Orange, blue, and green nodes represent word, sentence, and document nodes, respectively. The arrows on the edges indicate the current direction of information flow. First, on the left, words are used to aggregate sentence-level information, and the resulting sentence representations are then used to aggregate document-level information. Next, on the right, sentences are updated with the new document representations, and words are updated with the new sentence representations.}
\label{fig:2}
\end{figure}

Inspired by \citet{wang-etal-2020-heterogeneous}, we implement a bidirectional information propagation approach, as shown in Figure \ref{fig:2}. This approach allows information to flow from word-to-sentence-to-document and from document-to-sentence-to-word. 

After obtaining the document representation from the document encoder in Section \ref{sec:gat}, we update the sentence nodes using the updated document nodes and then update the word nodes using the updated sentence encodings. We further iteratively update the document nodes and sentence nodes. The information flow is bidirectional: in each iteration, it first moves from word-to-sentence-to-document (see Sections \ref{sentenceEncoder} and \ref{documentEncoder}), and then from document-to-sentence-to-word. For the $t^{th}$ iteration, the document-to-sentence-to-word update process can be represented as:
\begin{equation}
    \begin{gathered}
    \mathbf{U}_{\mathcal{D} \rightarrow \mathcal{S}}^{t+1} = \mathrm{GAT}(\mathbf{H}_{\mathcal{D}}^t) \\
    \mathbf{H}_{\mathcal{S}}^{t+1} = \mathrm{FFN}(\mathbf{U}_{\mathcal{D} \rightarrow \mathcal{S}}^{t+1}) \\
    \mathbf{U}_{\mathcal{S} \rightarrow w}^{t+1} = \mathrm{GAT}(\mathbf{H}_{\mathcal{S}}^{t+1}) \\
    \mathbf{H}_{w}^{t+1} = \mathrm{FFN}(\mathbf{U}_{\mathcal{S} \rightarrow w}^{t+1}),
    \end{gathered}
\end{equation}
where $\mathbf{H}^{0}_{\mathcal{D}}$ is initialized with $\tilde{\mathbf{h}}^{\mathcal{D}}$ in Equation \ref{eq:ffn} in the first iteration.

As illustrated in Figure \ref{fig:2}, word nodes can aggregate document-level information from sentences. For example, a word node with a high degree indicates frequent occurrences in multiple sentences, suggesting it is a keyword of the document. Sentence nodes with a higher concentration of important words are more likely to contain significant information, making them suitable for forming key sections. Bidirectional information propagation enables a comprehensive exchange of information across different hierarchical levels.

\subsection{Model Workflow}\label{sec:model_architecture}
Figure \ref{fig:model} shows an overview of the proposed model, which requires two forward (word-to-sentence-to-document) passes with a document-to-sentence-to-word update step in between them.

In the initial forward pass, RoBERTa word embeddings serve as the input, which are processed simultaneously by 
both the Dependency Tree Transformers (DTT) and Constituency Tree Transformers (CTT), 
followed by mean-pooling in the sentence encoder (see Equation \ref{equ:meanPoolSentence}). The GAT layer in the document encoder then computes the document representation 
using the sentence representations from the sentence encoder, marking the end of the first forward pass. 

After the first forward pass, the document-to-sentence-to-word update step is activated. This process begins with the document-to-sentence update, which updates 
$\mathbf{h}_d^{s_i}$
($\mathbf{h'}_d^{s_i}$) and $\mathbf{h}_c^{s_i}$ ($\mathbf{h'}_c^{s_i}$). 
Next, the sentence-to-word refinement step is applied twice: first to update the word embeddings based on $\mathbf{h'}_d^{s_i}$, and again using the updated $\mathbf{h'}_c^{s_i}$.

After the document-to-sentence-to-word update step, the second forward pass begins. This works over the pruned syntax trees and graph representation of sentences and document nodes from the first pass. This second pass mirrors the first with a slight modification: the sentence encoder now uses two sets of word embeddings. The CTT processes word embeddings updated by $\mathbf{h'}_c^{s_i}$, while DTT takes word embeddings updated by $\mathbf{h'}_d^{s_i}$. The following steps proceed similarly to the first pass, which generate refined sentence representations (i.e., $\mathbf{h''}_c^{s_i}$, and $\mathbf{h''}_d^{s_i}$) and ultimately, a refined document representation $\mathbf{H}_{\mathcal{D}}'$. 

\section{Experiment}
\subsection{Setup}
\paragraph{Datasets.} We evaluate  our proposed model on six common long document classification datasets: CMU BOOK Summary~\cite{bamman2013new}, ECtHR~\cite{chalkidis-etal-2021-paragraph}, Hyperpartisan~\cite{kiesel-etal-2019-semeval}, 20News~\cite{LANG1995331}, Amazon product reviews (AMZ)~\cite{10.1145/2872427.2883037}, and Essays\cite{pennebaker1999linguistic}. Following \citet{lu-etal-2023-efficient}, we randomly sample product reviews longer than 2048 words from the Book category for the AMZ dataset. The statistics of the datasets are summarized in Table \ref{tab:1}.

\begin{table}[t]
\resizebox{\columnwidth}{!}{
\begin{tabular}{ccccc}
\hline
\textbf{Dataset}       & \textbf{Type}                               & \textbf{\# of Classes} & \textbf{\# of Instances} & \textbf{\begin{tabular}[c]{@{}c@{}}Average \# of \\ Words per Document\end{tabular}} \\ \hline
\textbf{Hyperpartisan} & Binary                        & 2                          & 754,000                     &       745                                                                                   \\ \hline
\textbf{AMZ}           &  \multirow{2}{*}{Multi-class}                                           &                5            &  4850                            &    12,356                                                                                      \\
\textbf{20News}        &   & 20                         & 20,000                      &          369                                                                                \\ \hline
\textbf{BOOK}          & \multirow{3}{*}{Multi-label} &              227              & 16,559                      &      575                                                                                \\
\textbf{ECtHR}         &                                             &     33                       & 11,000                      &       5530                                                                                   \\
\textbf{Essays}        &                                             &  5                          &     1255                        &    660                                                                                      \\ \hline
\end{tabular}}
\caption{Details of the Long Document Classification Datasets.}
\label{tab:1}
\end{table}

\begin{table*}[ht]
\resizebox{\textwidth}{!}{
\begin{tabular}{c|cccccc}
\hline
\textbf{Models}        & \textbf{Hyperpartisan} & \textbf{20News} & \textbf{AMZ} & \textbf{BOOK} & \textbf{ECtHR} & \textbf{Essays} \\ \hline
\textbf{BERT w/ pre-training}~\cite{devlin-etal-2019-bert}          &    91.8           &    84.7    &  51.1   &   58.2   &   71.7    &   69.3     \\
\textbf{ToBERT}~\cite{9003958}           &      89.5         &   85.5     &  54.6   &   57.3   &   77.2    &    72.2    \\
\textbf{Longformer}~\cite{beltagy2020longformer}    &     93.7          &   83.4     &  56.4   &   58.5   &   81.5    &   74.4     \\
\textbf{BERT+Random}~\cite{park-etal-2022-efficient}   &      89.3         &    85.0    &   56.8  &   59.2   &   72.8    &    70.1    \\
\textbf{BERT+TextRank}~\cite{park-etal-2022-efficient} &      91.2         &    84.7    &  56.9   &  58.9    &   73.5    &   70.9     \\
\textbf{H3-pooler}~\cite{lu-etal-2023-efficient}     &     94.2         &   84.1     &  57.7   &   60.5   &   82.1    &    -    \\ \hline
\multirow{2}{*}{\textbf{Ours}}          &     \textbf{95.4}        &    \textbf{87.0}    &  \textbf{59.7}   &   \textbf{62.9} & \textbf{84.9}   &    \textbf{82.0}   
\\
~ & $\pm$0.92 &  $\pm$0.65 & $\pm$1.31 & $\pm$1.23 & $\pm$1.16 & $\pm$0.77
\\\hline
\end{tabular}}
\caption{Comparison to previous methods on the six long document classification datasets with pre-training. We use the reported scores from the original paper, except for the Essays dataset. Bold: best scores in each column.}
\label{tab:2}
\end{table*}

\paragraph{Implementation Details.} 
The model employs an initial learning rate of 0.1, which is subsequently reduced by 80\% in each iteration if the validation accuracy declines compared to the previous iteration. The batch size is 10. 
For the tree-transformers, the same hyper-parameter settings are used as in \citet{ahmed2019you}.
The statement encoding unit uses a GAT with six attention heads. For threshold (i.e., $\tau$) selection, we experiment with values ranging from 0.05 to 0.5 in intervals of 0.05. The best performance was found in the $\left[0.2,0.3\right]$ range. We then refine the experiments using 0.01 intervals, selecting the optimal threshold for each corpus: 0.21 for Hyperpartisan, 0.26 for 20News, 0.22 for AMZ, 0.24 for BOOK, 0.22 for ECtHR, and 0.27 for Essays. The model's parameters are trained using the ``Adagrad'' optimizer~\cite{lydia2019adagrad}.
The performance evaluation of our models has been conducted using 10-fold cross-validation. To facilitate this cross-validation process, we have utilized the StratifiedKFold function from the scikit-learn package. All experiments have been conducted in an Ubuntu 22.04 LTE environment, leveraging a 48GB NVIDIA RTX A6000 GPU. For parsing the sentences and generating the tree representations, we have used the Stanford Core-NLP parser \cite{coreNLP}. 

\subsection{Baseline Models}
Following \citet{lu-etal-2023-efficient}, we compare our methods with Transformer-based models with pre-training and a State-Space Model (SSM) system. 

\textbf{BERT with Pre-training} This simplest approach involves fine-tuning BERT~\cite{devlin-etal-2019-bert} after truncating long documents to the first 512 tokens. A fully connected layer is then applied to the [CLS] token for classification.

\textbf{ToBERT} Transformer over BERT (ToBERT) is a hierarchical approach designed to process documents of any lengths~\cite{9003958}. It divides long documents into chunks of 200 tokens and applies a Transformer layer to the BERT-based representations of these chunks.

\textbf{Longformer} is designed to handle longer input sequences with efficient self-attention that scales linearly with the sequence length, allowing it to process up to 4,096 tokens~\cite{beltagy2020longformer}.

\textbf{BERT+TextRank} To address BERT's 512-token limitation, \citet{park-etal-2022-efficient} augment the first 512 tokens with a second set of 512 tokens selected using TextRank \cite{mihalcea-tarau-2004-textrank}, an efficient unsupervised sentence ranking algorithm.

\textbf{BERT+Random} As an alternative method to BERT+TextRank, \citet{park-etal-2022-efficient} augment the first 512 tokens by selecting random sentences up to an additional 512 tokens.

\textbf{Hungry Hungry Hippo with Max Pooling (H3-pooler)}
H3 \cite{fu2023hungry} is an SSM-based method designed for simultaneous multi-object tracking, maintaining and updating object states based on observed data. \citet{lu-etal-2023-efficient} enhanced this model by inserting a max pooling layer between each SSM block, creating the H3-Pooler.

\section{Results and Discussions}
\subsection{Performance Comparision}

\begin{table*}[t]
\resizebox{\textwidth}{!}{
\begin{tabular}{cc|cccccc}
\hline
\multicolumn{2}{c|}{\textbf{Methods}}                                                          & \textbf{Hyperpartisan} & \textbf{20News} & \textbf{AMZ} & \textbf{BOOK} & \textbf{ECtHR} & \textbf{Essays} \\ \hline
\multicolumn{2}{c|}{\textbf{Full Model}}                                                       & 95.4                   & 87.0            & 59.7         & 62.9          & 84.9           & 82.0            \\ \hline
\multicolumn{1}{c|}{\multirow{3}{*}{\textbf{Tree Structure}}} & Removing CTT                   & 91.3                   & 83.6            & 54.8         & 58.8          & 81.8           & 78.9            \\
\multicolumn{1}{c|}{}                                         & Removing DTT                   & 90.9                   & 83.1            & 54.1         & 58.7          & 81.7           & 78.2            \\
\multicolumn{1}{c|}{}                                         & Removing Entire Tree Structure & 88.4                   & 78.5            & 47.3         & 52.4          & 77.3           & 73.7            \\ \hline
\multicolumn{2}{c|}{\textbf{Removing GAT}}                                                     & 88.8                   & 78.9            & 47.4         & 50.9          & 77.8           & 74.4            \\ \hline
\multicolumn{2}{c|}{\textbf{Removing the Bidirectional Propagation}}                                   & 87.9                   & 75.9            & 45.2         & 48.9          & 76.1           & 72.7            \\ \hline
\end{tabular}}
\caption{Ablation experiment results on the six long document classification datasets.}
\label{tab:3}
\end{table*}

We compare our proposed model against previous baseline models on various evaluation metrics, as shown in Table \ref{tab:2}. We report accuracy for binary and multi-class tasks (Hyperpartisan, 20News, and AMZ) and macro-F1 scores for the multi-label classification problems (BOOK, ECtHR, and Essays). We conducted 10 runs of experiments for each dataset and report the average performance. Each row in the table presents the performance of a specific method on each dataset, with the best score for each dataset highlighted. Standard deviation is also provided.

The results demonstrate that our model consistently achieves superior performance across all datasets. For binary classification (Hyperpartisan), our model outperforms existing methods by approximately 1.3\%. In multi-class classification tasks (20News and AMZ), our model shows a marked improvement, exceeding baseline accuracies by about 2\% to 4\%. In multi-label classification tasks (BOOK, ECtHR, and Essays), our model demonstrates a significant enhancement, with macro-F1 scores improving by approximately 3\% to 10\% over the best baseline models. These results underscore the robustness and effectiveness of our proposed model in handling the complexities of long document classification across different types of classification tasks.

\subsection{Ablation Studies}
We are interested in studying the effectiveness and robustness of our model by analyzing various components, such as the tree structure, graph structure, and bidirectional information propagation. To understand the impacts of these factors, we conduct controlled experiments with three different settings: (a) removing the tree structure, which is further divided into three sub-experiments: removing the CTT structure, removing the DTT structure, and replacing the entire tree structure with the [CLS] token; (b) replacing the GAT with a max-pooling layer; and (c) removing the bidirectional information propagation, where only the first forward pass is used. This allows us to evaluate the influence of each module individually, without interference from the others. The results are summarized in Table \ref{tab:3}.
\paragraph{Effectiveness of the Tree Structure.}
Table \ref{tab:3} demonstrates the crucial role of the tree structure in our model's performance. Removing the CTT component decreases performance notably, especially on multi-class classification datasets like 20News and AMZ, indicating CTT's importance in capturing document context at the phrase level. Similarly, excluding the DTT component leads to the reduction of the performance, with scores dropping by 4-6\% across various datasets, which underscores DTT's role in capturing inter-word relations while classifying long text. The most significant degradation occurs when the entire tree structure is replaced, especially in the multi-label classification tasks (BOOK, ECtHR, and Essays), indicating that hierarchical document modeling is important for maintaining the structural integrity and contextual coherence of the long documents. 
\paragraph{Effectiveness of the Graph Structure.}
As shown in Table \ref{tab:3}, removing the GAT module results in performance declines across all datasets, with a more pronounced impact on tasks requiring relational information, such as multi-label classification. This suggests that the GAT module is essential for capturing relationships between different elements within a document, which allows the model to leverage dependencies and interactions that are important for accurate classification. The attention mechanisms provided by the GAT module help focus on important features and connections, which are particularly important for the tasks involving complex relational data. 
\paragraph{Effectiveness of the Bidirectional Information Propagation.}
The bidirectional information propagation approach further enhances our model's effectiveness by fine-tuning feature representations, as indicated in Table \ref{tab:3}. Removing this module results in significant performance reductions across all datasets, with the most marked impact observed on the BOOK and AMZ datasets. This decline indicates that the bidirectional propagation plays a critical role in polishing the feature representations obtained from previous layers, ensuring accurate capture of subtle and important details. By refining these features, the module helps the model better distinguish between different classes, particularly in datasets with high variability and complexity.
\subsection{Key Section Identification for Long Document}
\begin{figure}[t]
\includegraphics[width=\columnwidth]{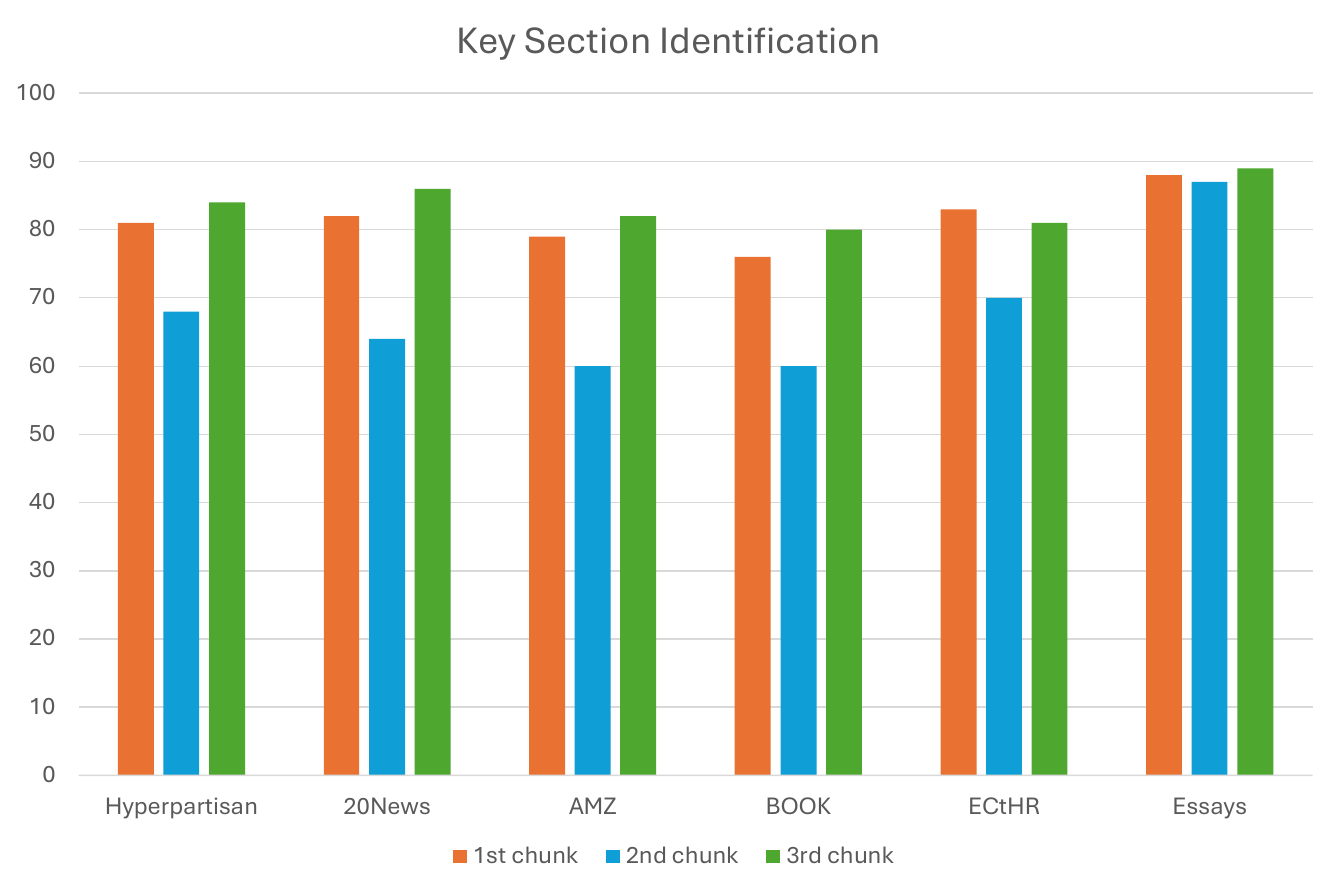}
\caption{Visualization of the percentage of sentences selected from different chunks across all datasets.}
\label{fig:3}
\end{figure}
In long document classification, not all sections are equally important. It is crucial to identify which parts of the documents contribute the most significant features for classification tasks. We conduct experiments by splitting each document into three chunks and calculating the number of sentences selected from each chunk. The results, shown in Figure \ref{fig:3}, indicate that the first and third chunks consistently contain more important information compared to the middle chunk across all datasets. 

Specifically, across all datasets except for Essays, approximately 76.6\% of sentences are selected from the first chunk, 83\% from the third chunk, and only about 62.6\% from the middle chunk. In contrast, the Essays dataset, being a questionnaire-style document, distributes important information more evenly, as each question provides different levels of information with equal importance. These observations highlight the importance of effectively capturing content at both the start and end of documents to enhance classification performance in long document tasks.

\section{Related Work}
Long documents present unique challenges. As the document length increases, maintaining context  becomes increasingly difficult, which makes the task substantially more complex compared to short text classification~\cite{9810897}. Early methods relied  on feature extraction techniques, 
where document length was not a significant issue. However, this changed with deep learning approaches using CNNs and RNNs that were implemented at different semantic levels, including character, word, and sentence levels~\cite{tang-etal-2015-document,yang-etal-2016-hierarchical}. CNNs often focus on local dependencies~\cite{gao-etal-2018-hierarchical,a11080109}, while RNNs~\cite{8675939,10.1145/3529836.3529935} are built for long-range dependencies. Transformer-based models, such as BERT~\cite{devlin-etal-2019-bert} and XLNet~\cite{10.5555/3454287.3454804}, have since come to dominate, but are limited by the number of tokens they can process, which poses a challenge for handling long documents. Methods for modifying transformer architectures to handle long documents include recurrent Transformers and sparse attention Transformers~\cite{dai-etal-2022-revisiting}. Transformer-XL~\cite{dai-etal-2019-transformer}, a recurrent approach, introduced a segment-level recurrence mechanism and a  positional encoding scheme, which enabled the model to capture long-term dependencies more effectively. Another approach, ERNIE-Doc~\cite{ding-etal-2021-ernie}, incorporated a continuous multi-segment attention mechanism and entity-aware pre-training  to capture comprehensive contextual information across longer texts. Alternatively, Sparse Transformers~\cite{child2019generating}  reduced the computational complexity of self-attention by selectively focusing on a subset of relevant tokens rather than all tokens in a sequence. Reformer~\cite{Kitaev2020Reformer:} then improved the efficiency of Transformers by using locality-sensitive hashing for sparse attention and reversible layers to reduce memory usage. Besides the aforementioned approaches, methods such as Longformer~\cite{beltagy2020longformer} and Big Bird~\cite{10.5555/3495724.3497174}  used a combination of local and global attention mechanisms to reduce computational complexity of standard self-attention. Recently, \citet{lu-etal-2023-efficient} used SSM to address the computational challenges (quadratic complexity in self-attention) caused by processing long sequences with traditional transformers. They introduced an SSM-pooler model, which incorporates a max pooling layer between each SSM block. This design allows the model to automatically extract important information from nearby inputs at each level and reduce the input length to half of that in the previous layer, which  significantly accelerates the speed of both training and inference. They compared their model with self-attention-based models and achieved comparable performance while being, on average, 36\% more efficient.
\section{Conclusion}
In this paper, we address challenges of long document classification by leveraging a novel graph-tree structure. By integrating syntax trees for sentence encodings and document graphs for comprehensive document encodings, our approach captures both fine-grained syntactic relationships and broader contextual dependencies. Using Tree Transformers and GAT ensures accurate modeling of hierarchical relationships within documents. Additionally, our bidirectional information propagation technique enhances the contextual representation, which enables a deeper understanding of content at all hierarchical levels. Notably, our approach not only overcomes the limitations of token constraints but also improves the performance and accuracy of long document classification, making it highly suitable for long document understanding. Potential extensions of this work could involve incorporating external knowledge through the integration of knowledge graphs. 
By linking document content with relevant external information, the model can further enhance its understanding and context-awareness and it would open new avenues for applications and improve the model's versatility across various domains.

\section*{Limitations}
Requiring two forward passes and parsing for the tree-structured transformers increases the time required  compared to the other models. This computational overhead should be taken into account when considering the deployment and scalability of the proposed models in practical applications. However, with label-wise attention cutoff values, some sentence and word nodes are pruned, which reduces the computational times significantly and the model takes similar time compared to the BERT-based model for document classification task.  Still, with some parallelization in the model implementation, the computational time can be reduced.
\section*{Ethics Statement}
We are using the publicly-available datasets, and we do not see any ethics issues in this paper.

\section*{Acknowledgements}
We thank all reviewers and area chairs for their constructive comments and feedback. This research is partially funded by The Natural Sciences and Engineering Research Council of Canada (NSERC) through a Discovery Grant to R. E. Mercer. F. Rudzicz is supported by a CIFAR Chair in AI.
\bibliography{anthology,custom}




\end{document}